# *SimpModeling*: Sketching Implicit Field to Guide Mesh Modeling for 3D Animalmorphic Head Design

ZHONGJIN LUO, SSE, The Chinese University of Hong Kong, Shenzhen, China
JIE ZHOU, School of Creative Media, City University of Hong Kong, China
HEMING ZHU, SRIBD, The Chinese University of Hong Kong, Shenzhen, China
DONG DU, GCL, University of Science and Technology of China, China
XIAOGUANG HAN*, SSE, The Chinese University of Hong Kong, Shenzhen, China
HONGBO FU*, School of Creative Media, City University of Hong Kong, China

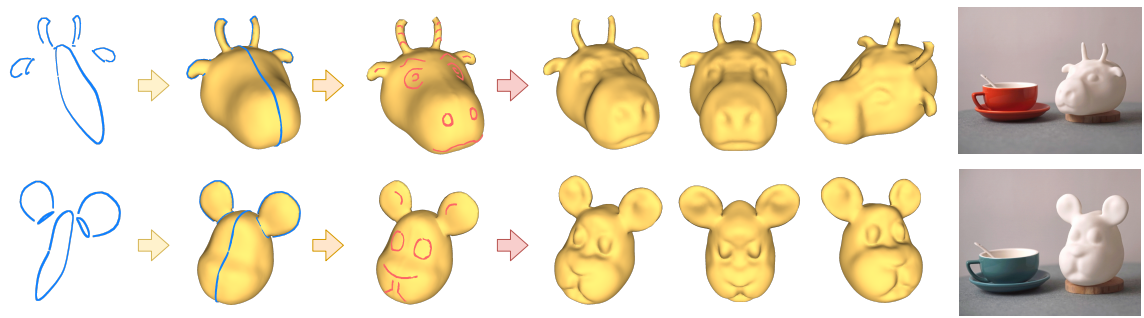

Fig. 1. We present *SimpModeling*, a novel sketching system designed for amateur users to create desired animalmorphic heads. It provides two stages for mesh modeling: coarse shape sketching where users may create coarse head models with 3D curve handles (blue), and geometric detail crafting where users may add geometric surface details by drawing sketches (red) on the coarse models. The two animalmorphic head models in this figure were created by a novice user without any 3D modeling experiences in ten minutes.

Head shapes play an important role in 3D character design. In this work, we propose *SimpModeling*, a novel sketch-based system for helping users, especially amateur users, easily model 3D animalmorphic heads - a prevalent kind of heads in character design. Although sketching provides an easy way to depict desired shapes, it is challenging to infer dense geometric information from sparse line drawings. Recently, deepnet-based approaches have been taken to address this challenge and try to produce rich geometric details from very few strokes. However, while such methods reduce users' workload, they would cause less controllability of target shapes. This is mainly due to the uncertainty of the neural prediction. Our system tackles this issue and provides good controllability from three aspects: 1) we separate coarse shape design and geometric detail specification into two stages and respectively provide different sketching means; 2) in coarse shape designing, sketches are used for both shape inference and geometric constraints to determine global geometry, and in geometric detail crafting, sketches are used for carving surface details; 3) in both stages, we use the advanced implicit-based shape inference methods, which have strong ability to handle the domain gap between freehand sketches and synthetic ones used for training. Experimental results confirm the effectiveness of our method and the usability of our interactive system. We also contribute to a dataset of high-quality 3D animal heads, which are manually created by artists.

---

*Corresponding authors

**ACM Reference Format:**
Zhongjin Luo, Jie Zhou, Heming Zhu, Dong Du, Xiaoguang Han, and Hongbo Fu. 2021. *SimpModeling*: Sketching Implicit Field to Guide Mesh Modeling for 3D Animalmorphic Head Design. In *UIST '21: 34th ACM Symposium on User Interface Software and Technology, October 10–13, 2021, Virtual.* ACM, New York, NY, USA, 16 pages. https://doi.org/10.1145/nnnnnnn.nnnnnnn

## 1 INTRODUCTION

Creating visually plausible 3D contents has always been a fundamental but laborious task in the modern game and film industries. Among the most prevalent 3D contents, articulated animalmorphic characters, specifically their heads, are highly demanded but challenging to make due to their highly diversified shapes and delicate surface details. Although various commercialized software packages, e.g., Maya, 3ds Max, and ZBrush, have been developed for modeling complex 3D shapes, they expect users to possess considerable expertise and proficiency in 3D modeling, thus hindering most novice users from instantiating their creative ideas.

Unlike traditional interactive shape modeling pipelines supported by most commercialized software, sketch-based modeling enables amateurs to get involved in 3D shape customization in a more intuitive fashion. After the seminal work of Teddy [14], in the past decades, researchers have extensively explored various interactive sketch-based modeling schemes, aiming to strike a good balance between the intuitiveness of modeling operations and the expressiveness of the resulting system. However, with existing tools, intensive and laborious interactions are still required to craft shapes with details, e.g., animalmorphic heads. With recent progress in deep learning, researchers resort to data-driven methods to simplify the interactive modeling process. While these learning-based systems, e.g., *SAniHead* [8], may generate visually plausible shapes with surface geometric details from a small number of strokes, they are limited by their fully learning-driven nature and may fall short in synthesizing novel shapes deviating from their small training datasets.

In this paper, we propose *SimpModeling*, a novel sketch-based modeling system for amateur users to create 3D animalmorphic head models with ease. To achieve great controllability without introducing excessive operations, we decompose the complex modeling task into two stages, i.e., **coarse shape sketching** and **geometric detail crafting**. In the first stage, users may guide the generation of coarse animalmorphic head models with a few 3D strokes. By seamlessly integrating the explicit mesh representation with implicit learning, this stage generates visually plausible animalmorphic heads respecting the input strokes. In the second stage, users may emboss the coarse head models with geometric details (e.g., wrinkles, noses, and eyes), by simply sketching on the model surfaces. The synthesis of geometric details from the on-surface sketch is achieved by optimizing a coarse mesh to fit a pixel-aligned implicit field derived from an integration of projected sketch images and rendered coarse depth maps. However, extracting a high-quality mesh from an implicit field is extremely time-consuming, which dramatically hinders its usage in real-time interaction systems. To address this issue, we propose to use the implicit function to guide the deformation of the coarse mesh, and the resulting system reaches a good balance between efficiency and output quality.

To train an expressive back-end model for our system, we constructed a data warehouse, which comprises 1,955 highly diversified animalmorphic head models (see Fig. 6), covering 17 categories, and annotated with rich 3D semantic shape contours. We conducted a user study, which shows that our proposed system strikes an good balance between controllability and simplicity. Our system can assist amateur users in crafting desired animalmorphic heads with diversified deformations and dedicated surface details by using a small number of strokes (see Fig. 1).

The major contributions of this work are summarized as follows:

- We design an easy-to-use and controllable sketch modeling interface for animalmorphic head modeling. In particular, it takes around ten minutes for novice users to create a desired animalmorphic head model using only a small number of 3D strokes.
- We propose a coarse-to-fine shape inference method, which seamlessly integrates explicit and implicit shape representations to guarantee reconstruction quality and efficiency.
- We contribute to the largest animalmorphic head dataset that consists of 1,955 high-quality animalmorphic head models. Each model in the proposed dataset is carved manually by artists and carefully annotated with 3D contour annotations. We will make the dataset together with the sketch modeling system publicly available to the research community.

## 2 RELATED WORK

This section presents relevant studies in sketch-based modeling. Most early solutions focus on optimizing 3D shapes to meet the constraints inducted from input sketches, by using handcrafted empirical principles [7, 18, 26]. With the emergence of deep learning techniques, deep neural networks have been extensively adopted for inferring 3D models from input sketches. 3D shape modeling can be roughly divided into two streams: articulate object (e.g., chairs, tables, lamps) design and organic shape creation. Among organic shapes, character shapes have received the majority of attention for their high demands. Even though character heads are inherently more diversified and contain more fine-grained details than other components, head modeling itself has not received much attention. Below we discuss character head modeling separately, since it is most closely relevant to our task.

*Character Head Modeling.* Some systems [14, 21, 23] originally targeted at modeling freeform shapes might be adopted for character head design. However, they might produce over-smooth or unrealistic results due to the lack of the domain knowledge of character heads. To address this issue, *DeepSketch2Face* [12] proposed an intuitive sketch modeling system for modeling human faces with diversified shapes and exaggerated facial expressions, but their system is confined to the parametric human face space. Since parametrizing animalmorphic heads into a low-dimensional parametric space is inherently difficult, *SAniHead* [8] poses a view-surface collaborative network to model animalmorphic head models via 3D template mesh deformation guided by 2D sketches. Although amateurs may create animalmorphic heads with *SAniHead* by using a few strokes, the fully-learning-based nature of *SAniHead* prohibits users from freely customizing shapes they expect (see comparisons in Fig. 9). In this paper, we integrate explicit mesh generation and implicit-based shape inference to reach a good balance between simplicity and controllability.

*Geometrical Sketch-based Modeling.* The research in sketch-based modeling can date back to the last century. As a seminal work of sketch-based modeling, Teddy [14] introduces a real-time, interactive method for constructing rotund shapes from input sketches. Follow-up works [2, 11, 16, 17, 25, 32] have focused on solving for interpolation functions defined by 2D sketches but tend to generating over-smooth shapes. To address this issue, several methods [20, 33, 40] utilize different types of strokes to form constraints to generate shapes with sharp features, at the cost of more laborious and less intuitive sketching process. Unlike the above works, which take 2D strokes as input, FiberMesh [23] aims to model freeform surfaces from a collection of 3D curves, which serve as handles for further shape editing. Inspired by FiberMesh, several well-designed 3D sketching systems [1, 27, 31] have been proposed for creating man-made object models. While the aforementioned 3D-sketch-based modeling systems significantly reduce the ambiguity of 2D-sketch-based modeling and well fit for modeling rough shapes, they are not capable for users to carve plausible surface details, such as facial organs of characters. Our proposed two-stage modeling approach enjoys the merits of both 3D sketching on controllable global shape generation and 2D sketching on intuitive detail crafting.

*Deep Sketch-based Modeling.* In the past few years, 3D deep learning has become a viable option for shape analysis and reconstruction. Inspired by the recent advance in 3D deep learning, Nishida et al. [24] proposed a CNN-based network to infer the parameters of urban building models from input sketches. Huang et al. [13] presented a similar model to learn progressive modeling of artificial objects. However, these methods only work well for specific shape categories that can be easily parameterized. To deal with general shapes, Delanoy et al. [6] adopted a volumetric representation for 3D objects to learn an end-to-end network to map input sketches to voxels. Nevertheless, their results are usually of low resolution due to the high memory consumption required for 3D convolutions. With the awareness of image-to-image learning's superiority, the works of [21, 22] generate depth maps together with normal maps from input sketches and fuse them to obtain complete 3D models. However, they expect users to draw very carefully to infer accurate depth and normal maps. Smirnov et al. [34] applied Coons patches to learn shape surfaces, but their method is limited to generating smooth shapes. Du et al. [9] decomposed 3D objects into parts and adopt implicit learning with low-resolution mesh extraction for distinct parts to guarantee accuracy. The method proposed by Li et al. [19] is also designed for CAD object modeling. None of the aforementioned works is targeted at animalmorphic head design.

The most relevant work to ours is *SAniHead* [8], which introduced a view-surface collaborative mesh generative network to enhance shape details. Although *SAniHead* can generate detailed animalmorphic heads by using a few strokes, it does not support accurate control of global shape. Recently, coarse-to-fine strategies have been widely adopted for 3D reconstruction and sketch-based modeling [12, 29, 30, 36]. For instance, *DeepSketch2Face* [12] consists an Initial Sketching Mode to create a coarse face and a Gesture-based Refinement Mode to refine the coarse shape. Since animalmorphic heads are highly diversified on global shapes and contain dedicated surface details, existing systems either require excessive human efforts or struggle to generate controllable outputs. To address this issue, we incorporate voxel-aligned and pixel-aligned implicit fields seamlessly into the coarse-to-fine interactive modeling process. The proposed system decomposes the modeling process of animalmorphic heads into two stages and strikes a good balance between controllability and simplicity, as detailed in the next section.

## 3　WORKFLOW

In this section, we will introduce the user interface and interaction design of our system. As illustrated in Fig. 2, our system decomposes the dedicated modeling process of animalmorphic heads into two stages: coarse shape sketching and geometric detail crafting. In Stage I, users can focus on molding a desired animalmorphic head's coarse appearance(e.g., ears, horns) with 3D curve drawing and manipulation tools. After finalizing the coarse shape, they may switch to Stage II, where they can freely carve fine-grained surface details by drawing on the surface of the coarse shape.

### 3.1　Coarse Shape Sketching

A user starts this stage by drawing a closed curve on the sketch pad for depicting a desired animalmorphic head profile contour. The system will then immediately generate a smoothed shape as an initial animalmorphic head mesh through interpolation. The user can use the provided **extrusion** tool to model head components like ears and horns: the user first draws a closed stroke on the mesh surface to indicate the region extruded, followed by an additional stroke describing a single-view silhouette. Moreover, we deliberately integrate the system with an **add-curve** tool and a **handle-deformation** tool, through which the user can freely manipulate the shape by setting and editing handles on the shape surface. The user may set a shape to be symmetric to get its mirrored part auto-completed by the system, since symmetric heads are often desired.

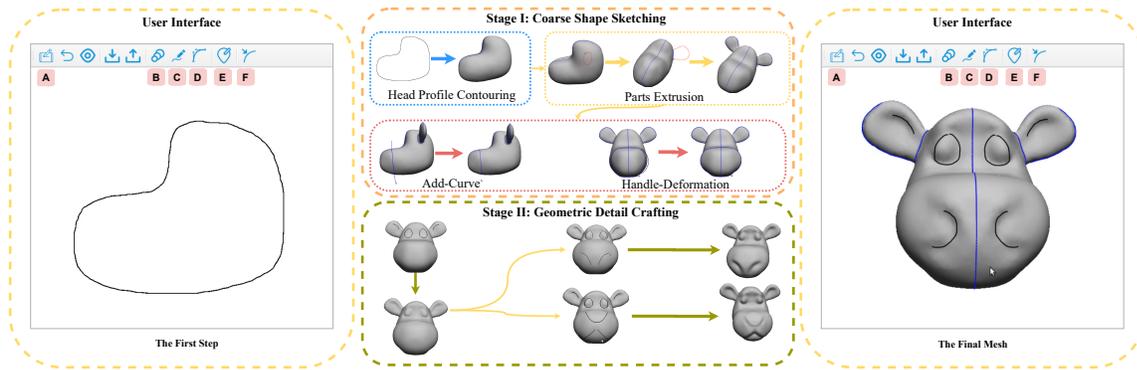

Fig. 2. Our system consists of two stages: Coarse Shape Sketching and Geometric Detail Crafting. In Stage I, users may create a coarse animalmorphic head by drawing an animalmorphic head contour. An **extrusion** tool is provided for creating animalmorphic head components like ears and horns. Our system also provides the **add-curve** and **handle-deformation** operations for further shape editing. In Stage II, users may depict surface details on the coarse animalmorphic head generated from Stage I, and our system automatically generates high-quality geometric surface details. The main tool bar is at the top of the user interface. From the left to right are **(A) clear canvas, (B) extrusion, (C) add-curve, (D) handle-deformation, (E) surface carving, and (F) smooth.**

### 3.2 Geometric Detail Crafting

In Stage I, users may customize diversified initial animalmorphic head meshes. However, the shapes generated from Stage I are often over-smoothed and lack expressive surface details. In Stage II, users may intuitively carve organic surface details, e.g., eyes, nose holes, and wrinkles. To do this, the user first hits the **surface carving** button (Fig. 2-E) and then draws feature curves (e.g., a closed round curve for an eye) on the coarse shape to depict the surface details. Our deep-learning-driven backend will immediately generate a high-quality animalmorphic head model with expected surface details after the user clicks the **inference** hot-key. Our system also supports auxiliary operations such as the **revert** and **smooth** operations. Furthermore, users are allowed to move back and forth between the two stages, and an option is provided for users to decide whether to keep the detail curves or not when they move back to the first stage from the second stage.

## 4 METHODOLOGY

In this section, we will introduce the back-end of our interactive sketch-based modeling system, i.e., the models and algorithms designed to parse users' inputs into high-quality animalmorphic models that respect the input sketches.

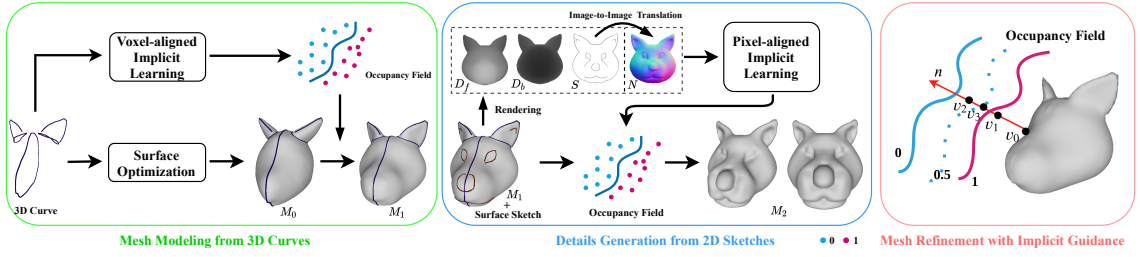

Fig. 3. The proposed two-stage mesh modeling can generate visually plausible shapes under the guidance of implicit functions. In Stage I, given 3D strokes as input, our system generates an initial mesh $M_0$ and learns a corresponding implicit field from the input. Then the coarse animalmorphic mesh $M_1$ is generated by deforming $M_0$ to fit the implicit field. In Stage II, our system first renders 2D sketch $S$ (from 3D strokes), the front and back view depth maps ($D_f$, $D_b$), and then predicts a normal map $N$ from the sketch with an image-to-image translation network. Afterward, all these four images ($D_f$, $D_b$, $S$, and $N$) are taken as input for the pixel-aligned implicit learning. The learned implicit field will guide the deformation of $M_1$ to produce a more detailed mesh $M_2$.

### 4.1 Mesh Modeling from 3D Curves

As mentioned in Sec. 3, our system allows users to customize and manipulate shapes with a few 3D curves. However, modeling complex shapes with sparse 3D curve guidance is inherently an ill-posed problem. The pioneering work of Nealen et al. [23] tried to address this issue by optimizing a smooth surface that interpolates 3D control curves. We argue that the strong prior knowledge learned from the animalmorphic dataset would significantly speed up the interactive coarse shape modeling process by providing data-driven interpolations among 3D control curves. Therefore, we propose to adopt a learned implicit-based deep model to infer a better coarse shape. As it is time-consuming to extract a high-fidelity mesh directly from the implicit field, we turn to use the learned implicit-based deep model to guide the refinement of an initially interpolated fair surface.

*Initial Mesh Generation.* A pillow-like fair surface is first generated so that its silhouette from the side is seamlessly aligned with a user-specified closed curve. Specifically, given a user-specified closed curve depicting the silhouette of a desired animalmorphic head from the side, we perform *Constrained Delaunay triangulation* [3] to fill the area and apply a symmetry constraint to generate a watertight mesh. Inspired by FiberMesh [23], we then apply predefined Laplacian Magnitudes (LMs) to the user-specified initial curve and diffuse the LMs to the rest of the mesh by solving a least-squares minimization problem:

$$\arg\min_{\{m_i\}} \left\{ \sum_i \|L(m_i)\|^2 + \sum_{c \in C} \|m_c - m'_c\|^2 \right\}, \tag{1}$$

where $L$ denotes the Laplace-Beltrami operator [35], $\{m_i\}$ denotes the LMs on all mesh vertices, and $\{m_c\}$ denotes the LMs on the initial user-specified curve. After acquiring LMs $\{m_i\}$ on mesh vertices, the target Laplacian [23, 38] $\delta_i$ for vertex $\mathbf{v}_i$ is calculated with $\delta_i = A_i \cdot m_i \cdot \mathbf{n}_i$, where $A_i$ is the area estimate of $\mathbf{v}_i$ and $\mathbf{n}_i$ is the normal of $\mathbf{v}_i$. Finally, we update the positions of the vertices by solving another least-squares minimization problem:

$$\arg\min_{\{\mathbf{v}_i\}} \left\{ \sum_i \|L(\mathbf{v}_i) - \delta_i\|^2 + \sum_{c \in C} \|\mathbf{v}_c - \mathbf{v}'_c\|^2 \right\}, \tag{2}$$

where the first term forces the vertex Laplacians to be close to the integrated target Laplacians, and the second term penalizes the vertices drifting away from the sketch stroke. It is worth mentioning that, unlike FiberMesh [23], which solves the Laplacian optimization (Equa. 1 and Equa. 2) whenever the user draws a new 3D stroke, we solve Equa. 1 and Equa. 2 only when the user specifies the initial silhouette. The subsequent sketching will only trigger the implicit-guided mesh refinement.

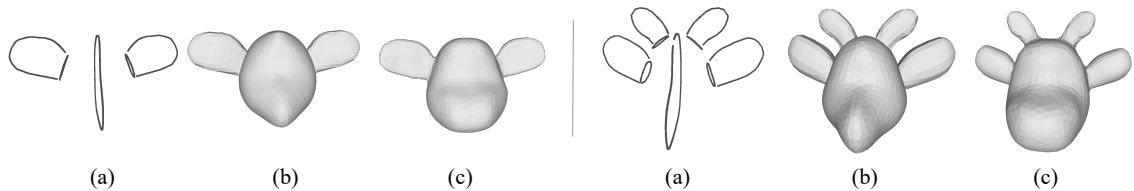

Fig. 4. (a) input 3D curves, (b) models generated by FiberMesh, (c) coarse models by our method. Compared with the results of FiberMesh, our models are closer to animal heads.

*Mesh Refinement with Voxel-aligned Implicit Guidance.* As illustrated in Fig. 4, we argue that a more intelligently interpolated coarse shape would potentially facilitate the interactive modeling process. To achieve this, we propose to adopt a continuous occupancy function $f(z, p) : Z \times \mathbb{R}^3 \mapsto [0, 1]$ to refine the initial mesh generated by interpolation and optimization, where $Z$ denotes the multi-resolution geometrical features [4] extracted from voxelized sketch strokes. For each query point in space, the occupancy function $f(z, p)$ consumes the sketch latent code $z$ and its coordinate $p$ to predict whether the point is inside or outside the shape.

Fitting the initial mesh directly to the target mesh extracted from the occupancy field $f(z, p)$ with non-rigid ICP could be a straightforward solution. However, extracting a high-resolution target mesh from the implicit field is time-consuming and thus not suitable for interactive systems. Therefore, instead of extracting a target mesh directly from the implicit field, we propose to refine the initial mesh progressively under the guidance of the implicit field, for all vertices $\{\mathbf{v}_i\}$ on the initial mesh $M_0$:

$$\mathbf{v}_i = \underset{\{\mathbf{v}_i\}}{\arg\min} \left\{ \sum_i \lambda \cdot \|L(\mathbf{v}_i)\|^2 + \sum_i \|\mathbf{v}_i - \mathbf{v}'_i\|^2 \right\}, \quad (3)$$

where $\mathbf{v}'_i$ denotes the $i$-th vertex's target position, and $\lambda$ is to balance the smoothness and fitting accuracy. Following the idea of ICP, we iteratively update the vertex positions $\{\mathbf{v}_i^k\}$ of the initial mesh to fit the 0.5 valued surface of the implicit field:

$$\mathbf{v}_i^{k+1} = \mathbf{v}_i^k - d_i^k \cdot sign(f(\mathbf{z}_i^k, \mathbf{v}_i^k) - \alpha) \cdot \mathbf{n}_i^k, \quad (4)$$

where $d_i^k$ is the displacement step length of $\mathbf{v}_i^k$ along the normal direction at the $k$-th iteration, $sign(\cdot)$ denotes the sign function, and $\alpha$ denotes the occupancy threshold (set to 0.5 by default). The step length $d_i^k$ is 0.1 at the beginning and decreases with a ratio of 0.5 whenever the sign of occupancy changes (i.e., when the explicit mesh vertex penetrates the implicit surface). In our system, the number of iteration is set to five empirically to reach a balance between efficiency and reconstruction quality. Eventually, as shown in Fig. 3, we can obtain a refined coarse animalmorphic head $M_1$ from 3D curve inputs in real-time.

*Handle-based Mesh Deformation.* Our system allows users to specify control curves on the model surface. The specified control curves serve as handles for manipulating the geometry. Every time when the user specifies such a control curve, a sparse linear matrix is established and then pre-decomposed in a background session to speed up the subsequent computation of handle-based deformation. Moreover, our system allows users to specify the target shape of control curves through sketching (see the demo video). Soon after the target shape of a control curve is specified, a handle-based Laplacian deformation [35] is performed to generate the expected shape.

### 4.2 Details Generation from 2D Sketches

In the previous stage, the user can easily create diversified animalmorphic heads $M_1$ from a few 3D strokes with implicit-guided handle-based deformations. However, the animalmorphic heads generated with sketch strokes often miss important surface details, e.g., eyes, wrinkles, and mouth. To address this, our system allows users to emboss the animalmorphic head with surface geometric details through directly sketching on the shape surface. This feature is powered by a pixel-aligned implicit function.

*Pixel-aligned Implicit Learning.* To enrich the coarse shape with surface geometric details, we turn to pixel-aligned local features [29] for detailed implicit shape learning. To keep the global shape of the coarse shape, we render $M_1$ using suggestive contours [5] and integrate the rendered contours together with the user-specified sketch on $M_1$ into a $256 \times 256$ sketch image, denoted as $S$ (see Fig. 3), which is used as part of the input for implicit function learning.

However, compared with real photo images, the resulting sketch images contain extremely sparse information. Therefore, the problem of constructing a detailed implicit field conditioned on the sparse sketch input is highly underdetermined. Since the normal maps reflect the relative variation of the shape, they may provide intuitive and continuous guidance for implicit learning. Inspired by [39], we use an image-to-image translation network [15] to predict a normal map $N$ from the sketch image $S$ as the intermediate guidance. An implicit function, which is conditioned on pixel-aligned local features extracted from $N$ and $S$, is then adopted to map the positions in space to occupancy values.

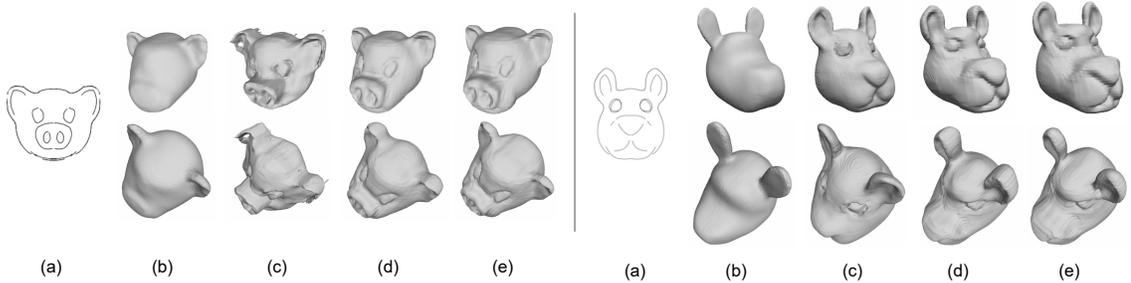

Fig. 5. A brief ablation study of the depth guidance to reduce depth ambiguity. Each input sketch (a) is followed by the coarse mesh generated in Stage I (b), the results without depth guidance (c), the results with front view depth as guidance (d), and the results generated by the full model with front and back view depth (e).

*Ambiguity Reduction with Depth Guidance.* Pixel-aligned implicit learning could generate shapes with visually plausible appearances from the front view but might lead to inaccurate prediction along the depth axis. This is caused by the ambiguity introduced when lifting pixel-aligned 2D features to 3D shapes. To address this issue, apart from $S$ and $N$, we additionally render the front and back depth maps of the coarse animalmorphic head $M_1$ as the input for the pixel-aligned implicit function. As illustrated in Fig. 5, the quality of the shape is significantly improved with the additional guidance of the front and back depth maps.

| SAniHead | | | | | | | | Ours | | | | | | | | |
|---|---|---|---|---|---|---|---|---|---|---|---|---|---|---|---|---|
| Sheep | 18 | Bear | 9 | Camel | 3 | Cat | 10 | Cow | 18 | Sheep | 115 | Bear | 120 | Camel | 72 | Cat | 75 | Cow | 130 |
| Deer | 18 | Dog | 19 | Wolf | 14 | Hippo | 7 | Horse | 18 | Deer | 175 | Dog | 140 | Wolf | 150 | Hippo | 160 | Horse | 84 |
| Mouse | 5 | Pig | 10 | Lion | 35 | Rabbit | 3 | Rhinoceros | 15 | Mouse | 93 | Pig | 69 | Lion | 175 | Rabbit | 160 | Fox | 60 |
| Orangutan | 1 | Kangaroo | 2 | Elephant | 5 | Dinosaur | 10 | | | Tiger | 105 | Monkey | 72 | | | | | | |
| 220 models, 19 categories | | | | | | | | | | 1955 models, 17 categories | | | | | | | | | |

Table 1. Dataset statistics for *SAniHead* and our proposed *3DAnimalHead*. *3DAnimalHead* is one scale larger than *SAniHead*, and enjoys much better data balance across categories.

*Mesh Refinement with Implicit Guidance.* Similar to Section 4.1, the learned pixel-aligned implicit field is utilized to guide the iterative shape refinement. Specifically, whenever a user draws a stroke on the shape surface, the neighboring patches around the stroke are immediately treated as the target regions to be refined. We then densify the triangles of the target regions using mesh subdivision to achieve more expressive and stable deformation. Note that in this step, we apply Equa. 3 only to the target patches and set $\lambda = 0.2$ to preserve more geometric details. In addition, after refinement, we apply the bilateral normal filtering [41] to the regions surrounding the target regions to suppress boundary artifacts.

## 5 EVALUATION

### 5.1 Dataset Construction

The recent advance of deep learning-based methods relies heavily on the availability of large-scale datasets. While various human character datasets are publicly available, few datasets address 3D animalmorphic shapes. SMAL [42] builds a 3D animal dataset from animal toy figurine scans. However, this dataset is relatively small and contains only 41 animal models. *SAniHead* [8] introduces a dataset containing 220 3D animalmorphic head models from 19 categories. Since each category only contains a limited number of models, this dataset suffers from severe data imbalance among categories. To address this issue, we propose *3DAnimalHead*, the largest animalmorphic head dataset by far. Our dataset contains 1,955 high-quality animalmorphic head models manually created by artists, covering 17 common categories (see Table 1). Each animalmorphic head is carefully annotated with contour curves to facilitate future research on 3D animalmorphic head generation.

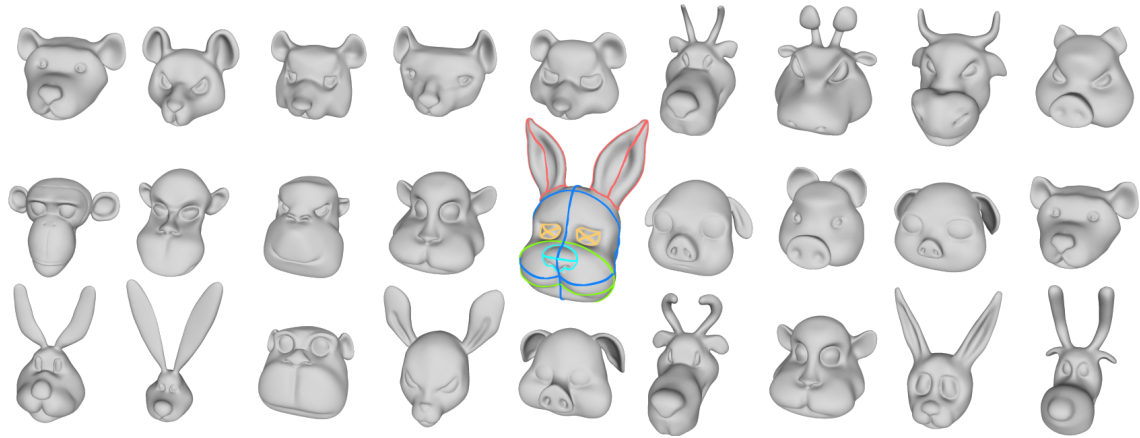

Fig. 6. The animalmorphic head models sampled from *3DAnimalHead*. Feature annotations are illustrated on a representative mesh with colored 3D curves.

*Animal Head Modeling.* To construct an expressive animalmorphic head dataset, we firstly collected 220 animal head images with diversified appearances, which belongs to 17 non-intersecting categories, e.g., cat, dog, pig, horse, etc (see the example reference images in Fig. 7 (a)). Specifically, to address the data imbalance issue, we manually balanced the number of figures belonging to each category. Two professional 3D modelers were hired to model 220 heads to be as similar as possible to the reference images. To further enhance the expressiveness and shape diversity of *3DAnimalHead*, we asked the modelers to exaggerate these 220 animalmorphic head models with the deformation tools in ZBrush (adjusting the global shape, ear position, expression, etc). With the augmentation of shape, 1,955 high-quality animalmorphic head models were generated (see the representative models in Fig. 7 (a)). It took about 20 minutes on average for an artist to model a head model and around 8 minutes for them to create an exaggerated model. We paid the artists $10 for each basic model and $1 for each exaggerated model on average.

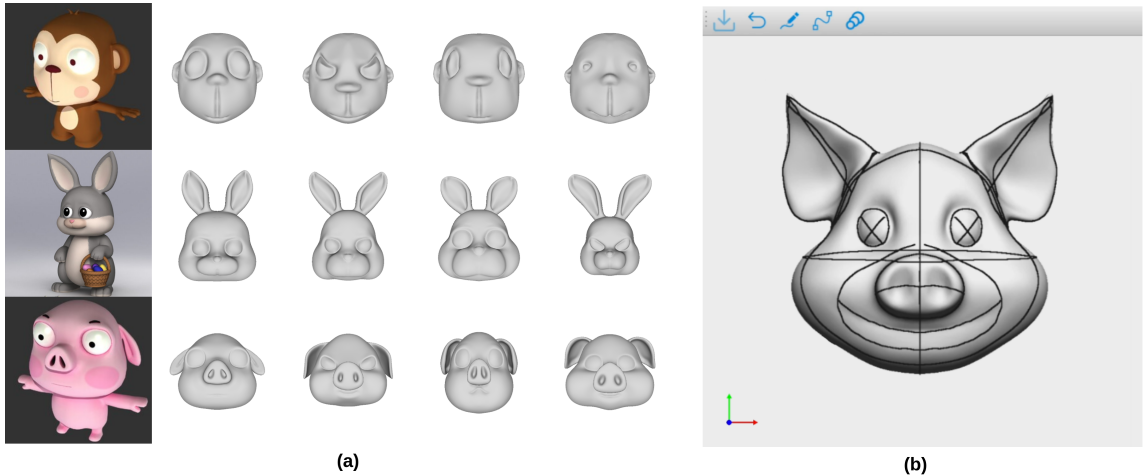

Fig. 7. (a) The first column is the reference images, the second column is the corresponding basic models, and the rest are the exaggerated models. (b) Our anotation interface.

*Animal Head Annotation.* To facilitate further research in animalmorphic head reconstruction and reasoning, each animalmorphic head model in *3DAnimalHead* is richly annotated with category labels and salient 3D contour curves. We developed an annotation tool and hired 8 annotators to annotate the contour curves (see Fig. 7) (b) for the annonation interface. In particular, such contour curves correspond to the most salient shape features of animalmorphic heads, e.g., head contours, ears, horns, which may provide strong guidance for shape reconstruction. The contours of different parts on animalmorphic head meshes are manually annotated using colored strokes. Fig. 6 illustrates sampled animalmorphic models from *3DAnimalHead*, with the corresponding 3D curves shown on top of a representative model.

### 5.2 Implementation Details

Our prototype system runs on a personal computer and supports either a mouse or a pen tablet as input. The user interface is implemented with the QT framework. CGAL [10] is adopted for 3D geometry processing and Eigen is utilized for solving optimization equations. The deep neural network models are implemented with PyTorch [28]. As shown in the accompanying video, running on a personal computer with Intel i7-7700 CPU, 16GB RAM, and a single Nvidia GTX 2080Ti GPU, our system supports real-time feedback.

*Network Architectures.* Inspired by IF-Net [4] and Pifu [29], we adopt a voxel-aligned implicit network to model coarse shapes, and a pixel-aligned implicit network to provide local geometrical details: **a) Voxel-aligned implicit network**. We firstly voxelize input 3D strokes into $128^3$ volumetric grids and adopt 3D convolutions as the encoder to obtain multi-scale feature maps ($128 \times 128 \times 128$, $64 \times 64 \times 64$, $32 \times 32 \times 32$, $16 \times 16 \times 16$, $1 \times 1 \times 1$). For each queried position, we fetch the features on the corresponding positions of the multi-scale feature maps, and concatenate them as the voxel-aligned positional features. A multi-layer perceptron is then applied to infer the occupancy from the positional features. **b) Pixel-aligned implicit network**. The sketch image, depth maps from the front and back views, together with the normal map, are scaled to $256 \times 256$ and concatenated channel-wise as the input of the pixel-aligned implicit network. A stacked hourglass network is then adopted as the backbone to extract image features. Given an arbitrary position in space, its pixel-aligned feature can be fetched from the concatenated feature maps through orthogonal projection. Finally, the space point's occupancy value can be predicted with a 6-layer MLP from the pixel-aligned features.

*Training Details.* The whole *3DAnimalHead* dataset is split into a training set of 1,900 models and a testing set of 55 models. We train the **voxel-aligned implicit network** and the **pixel-aligned implicit network** separately.

To train the **voxel-aligned implicit network**, we randomly sample 3,000 points from the input 3D sketch and then voxelize these points into $128^3$ 0-1 volumetric grids as the input of the network. For augmentation, we randomly drop stroke segments from the input 3D curves, which are used for mimicking users' inputs. It is worth mentioning that we remove the strokes of eyes, noses, and mouths as they are not relevant to the generation of coarse animalmorphic heads. We sample 60,000 points for each model in *3DAnimalHead* with the mixture of uniformed sampling and importance sampling near the animalmorphic heads' surfaces (1 : 9). The cross-entropy loss is adopted to measure the difference between predicted occupancy and the ground truth. We train the **voxel-aligned implicit network** with a batch size of 4 and a learning rate of $3 \times 10^{-4}$ for 100 epochs using the Adam optimizer.

To train the **pixel-aligned implicit network**, which infers a detailed shape from an input sketch, we firstly render the detailed 2D sketches of animalmorphic head models with suggestive contour [5]. The corresponding normal maps are predicted by a pre-trained pix2pixHD [37] from detailed input sketches. Since there are no ground-truth coarse shapes corresponding to the models in *3DAnimalHead*, we smooth the models in *3DAnimalHead* and render them to get the coarse depth maps. Silently different from the sampling scheme used for training the **voxel-aligned implicit network**, we sample 8,000 points for each model in *3DAnimalHead* with the mixture of uniformed sampling and importance sampling near the animalmorphic heads' surface (1 : 7). We train the pixel-aligned implicit network with a batch size of 2 and a learning rate of $2 \times 10^{-4}$ for 30 epochs with the Adam optimizer. And we use the MSE loss for network training.

### 5.3 User Study

To evaluate the usability and effectiveness of our system, and to identify any limitations or opportunities for future advancements, we conducted several user studies, including *Usability Study*, *Comparison Study* and *Perceptive Evaluation Study*.

*Training.* We recruited 15 subjects (8 males and 7 females, aged from 18 to 30, P1-P15) from different departments in a local university, including Social Science (5), Economics (3), Computer Science (4), and Life Science (3), to evaluate our system. Based on the pre-study survey, all of the participants had limited or even no experience in 3D modeling. At the beginning of the user study, each participant was given a short demonstration (around 8 minutes) showing the basic operations of our system. After the demonstration, each user had 15 minutes to get familiar with the tools provided in our system.

*Usability Study.* Since our proposed system is designed to assist users in instantiating their ideas, in the usability study, we asked the participants to create whatever they wanted with our system within 30 minutes and did not ask them to create a specific number of models. We observed that each participant created on average 3 to 4 models within the given time. Fig. 8 shows representative models created by the subjects with our system. It can be seen from this figure that our system supports users with limited knowledge in geometrical modeling to create animalmorphic heads with diversified shapes and rich geometric details. The system received positive feedback from most of the subjects, and they were deeply impressed by the intuitiveness, simplicity, and controllability of our system. **"I can hardly believe 3D modeling could be that easy before working with this system, since it simply generates whatever I want."** (P1) **"The tools provided are easy to use."** (p8) Some participants left some suggestions for our system. **"Some types of animal heads are difficult to create, such as elephants."** (P6) **"Drawing too dense may lead to unexpected results."** (P9)

By the end of the user study, we asked the subjects to fill in a System Usability Scale (SUS) questionnaire and a NASA Task Load Index (NASA-TLX) questionnaire to evaluate the usability and workload of our system. Fig. 10(a) illustrates the average score for each question in the SUS questionnaire. The subjects gave high ratings for all of the questions (87.7 out of 100.0 on average), confirming the great usability of our system. The results for NASA-TLX are also positive: the mental demand, physical demand, temporal demand, effort and frustration are at an extremely low level. Meanwhile, the performance score is relatively high indicating that our system allows users to customize desirable animalmorphic head models effortlessly.

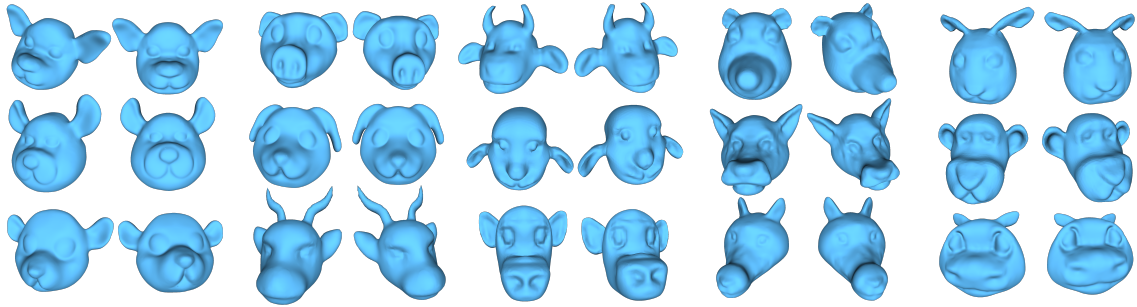

Fig. 8. The gallery of the representative models created using our system by 15 invited amateur users.

*Comparison Study.* After thoroughly reviewing the existing open-source sketch-based modeling systems, we choose FiberMesh [23] and SAniHead [8], which are most relevant to our work, for comparison. Before the formal user study, all the subjects were given an 8-minute tutorial and 15 minutes to get familiar with each system. Each subject was then given different reference images and asked to create 3D animalmorphic heads according to the reference image with the systems above (our system, FiberMesh, and SAniHead) in a randomized order. For each system, the participants were given at most 15 minutes. Note that the created 3D model was not required to strictly follow the reference image and differences were allowed. Fig. 9 shows the animalmorphic head models created using the compared three systems. Obviously, given the limited time, the participants managed to model more visually appealing shapes and freely craft vivid details as they desired with our system. The global shapes and the surface details created with our system are more in line with the referencing images. Moreover, we observed that most subjects got stuck in the global shape modeling process with FiberMesh. Some even ran out of time before crafting any fine-grained details. Most of the subjects successfully customized a detailed animalmorphic head with SAniHead, but some of them complained that they could not create surface details as they desired. It is worth mentioning that, although the predicted initial global shape sometimes did not fully meet subjects' expectations, they could easily adjust the shapes and depict surface details with the tools provided in our system.

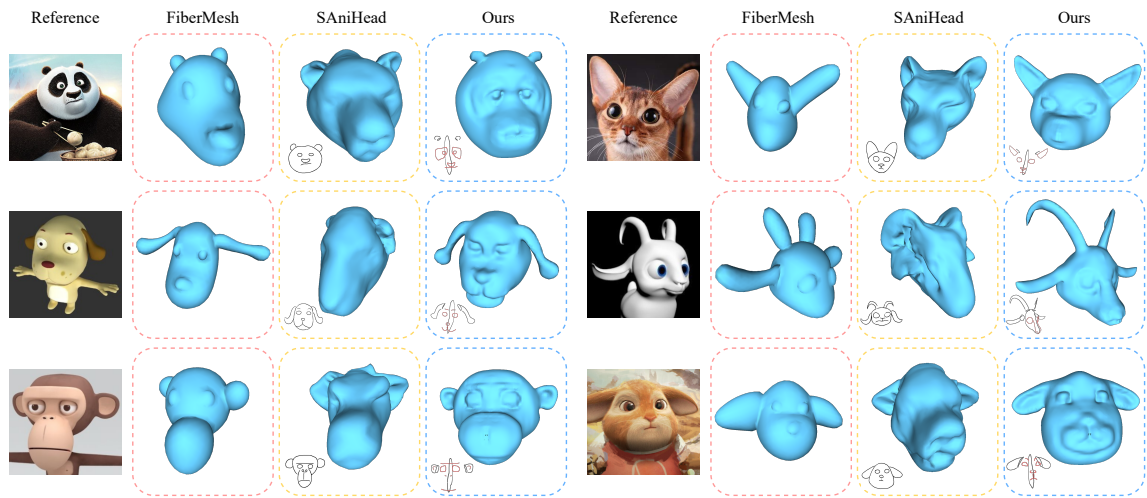

Fig. 9. The models created in the comparison study. Each reference image is followed by the models created with FiberMesh[23], SAniHead[8], and our system by the participants within 15 minutes.

*Perceptive Evaluation Study.* We further conducted a perceptive evaluation study to evaluate the quality of animalmorphic heads modeled using the three compared systems. In this study, we invited 60 subjects (30 male, 30 female, aged from 20 to 32), none of whom had participated in the previous two studies. For each subject, we randomly selected eight animalmorphic heads modeled by the amateurs from the *Comparison Study*, and asked them to score each model based on model quality and detail richness (with 1 denoting the lowest quality and 10 the highest quality). Fig. 10(c) shows the results for this study. As seen from Fig. 10(c), the animalmorphic heads modeled with our system received significantly higher marks than the counterpart systems, implying that our system could assist novice users in creating more realistic animalmorphic heads with rich details.

## 6 CONCLUSION AND LIMITATION

In this paper, we have presented *SimpModeling*, a novel sketch-based modeling system for amateurs to model 3D animalmorphic head models. To strike a good balance between simplicity and controllability, we decompose the modeling of animalmorphic heads into two stages: coarse shape sketching and geometric detail crafting. In Stage I, users can model the coarse shapes of animalmorphic heads through 3D sketch curve drawing. In the second stage, users may carve desired surface details through sketching on the surface of a coarse shape from Stage I. Both the stages are powered by novel hybrid back-end models, which integrate the prevalent implicit learning techniques with robust mesh deformation methods. The user study demonstrated that our system could help users efficiently create high-quality animalmorphic head models as desired.

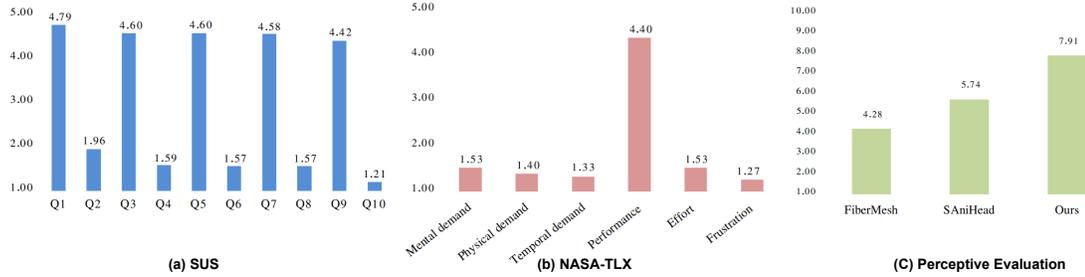

Fig. 10. (a) Mean scores of SUS in a 5-point scale. For the questions with the odd numbers, the higher the scores are the better; for the rest of the questions, the lower the scores are the better. (b) Mean scores of NASA-TLX in a 5-point scale. (c) Mean scores of *Perceptive Evaluation* in a 10-point scale.

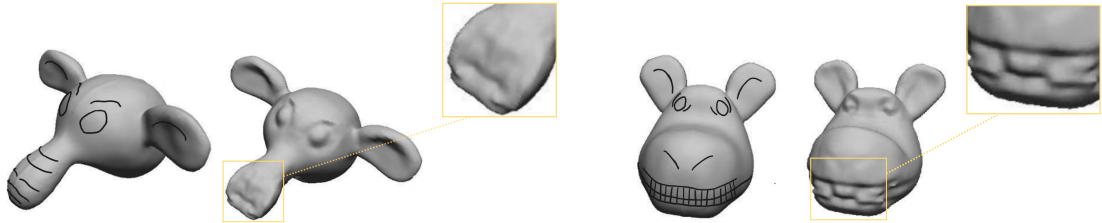

Fig. 11. The failure cases of our system when facing models in unseen categories and over-complicated details.

*Limitations and Future works.* Our current system suffers from the following limitations: 1) as shown in Fig. 11(Left), although our system is capable of modeling coarse animalmorphic heads from unseen categories, it might fail to generate some out-of-scope geometrical details. 2) as seen in Fig. 11(right), our model might fail to generate desired details when the detail sketch strokes are too dense, partly due to the limited input resolution for pixel-aligned implicit function and insufficient mesh resolution. 3) our current system cannot be directly extended to model shapes with complicated topologies, e.g., chairs, tables. In the future, we will extend our system to model detailed shapes from other categories, such as cartoon human bodies. A potential solution is to enlarge our dataset by introducing new categories and more detailed models under different categories. We will also incorporate part-based modeling into our sketch-based modeling pipeline to support the modeling of shapes with complicated topologies. We believe it would also be an intriguing future avenue to explore the possibilities to carve extremely-high resolution shape interactively. Meanwhile, we will make our system publicly available and collect more comments from users in more diversified background groups to further improve our system.

## 7 ACKNOWLEDGMENTS

The work was supported in part by Shenzhen Basic Research General Program under Grant JCYJ20190814112007258, NSFC-61902334, grants from the Research Grants Council of the Hong Kong Special Administrative Region, China